\documentclass{article}
 \usepackage[preprint]{template}

\usepackage[utf8]{inputenc} 
\usepackage[T1]{fontenc}    
\usepackage{hyperref}       
\usepackage{url}            
\usepackage{booktabs}       
\usepackage{amsfonts}       
\usepackage{nicefrac}       
\usepackage{microtype}      
\usepackage{xcolor}         
\usepackage{colortbl}
\usepackage{makecell}
\usepackage{algorithm}
\usepackage{algpseudocode}
\usepackage{amsmath}
\usepackage{xcolor}
\usepackage{multirow}
\usepackage{graphicx}
\usepackage{wrapfig}    
\usepackage{adjustbox}  
\usepackage{makecell}   
\usepackage{subcaption} 
\usepackage{wrapfig}  

\definecolor{lightgray}{gray}{0.92}
\newcommand{\thickhline}{\Xhline{1.2pt}}

\newcommand{\pub}[1]{{\color{gray}{\scriptsize{[{#1}]}}}}

\title{Towards Unified Modeling in Federated Multi-Task Learning via Subspace Decoupling}

\author{%
Yipan Wei$^{1}$ \quad Yuchen Zou$^{2}$ \quad Yapeng Li$^{1\dagger}$ \quad Bo Du$^{1\dagger}$\\
$^{1}$ School of Computer Science, Wuhan University, Wuhan, China. \\
$^{2}$ Faculty of Artificial Intelligence in Education, Central China Normal University, Wuhan, China. \\
\texttt{\{yipanwei,yapengli,bodu\}@whu.edu.cn}\\
\texttt{\{janemo\}@mails.ccnu.edu.cn}\\
{\small $^\dagger$ Corresponding author.}
}

\begin{document}
\maketitle

\begin{abstract}
Federated Multi-Task Learning (FMTL) enables multiple clients performing heterogeneous tasks without exchanging their local data, offering broad potential for privacy preserving multi-task collaboration. However, most existing methods focus on building personalized models for each client and unable to support the aggregation of multiple heterogeneous tasks into a unified model. As a result, in real-world scenarios where task objectives, label spaces, and optimization paths vary significantly, conventional FMTL methods struggle to achieve effective joint training. To address this challenge, we propose FedDEA (Federated Decoupled Aggregation), an update-structure-aware aggregation method specifically designed for multi-task model integration. Our method dynamically identifies task-relevant dimensions based on the response strength of local updates and enhances their optimization effectiveness through rescaling. This mechanism effectively suppresses cross-task interference and enables task-level decoupled aggregation within a unified global model. FedDEA does not rely on task labels or architectural modifications, making it broadly applicable and deployment-friendly. Experimental results demonstrate that it can be easily integrated into various mainstream federated optimization algorithms and consistently delivers significant overall performance improvements on widely used NYUD-V2 and PASCAL-Context. These results validate the robustness and generalization capabilities of FedDEA under highly heterogeneous task settings.

\end{abstract}

\section{Introduction}

Federated Learning (FL)~\cite{ fedavg2017,Industry_2019,Nature2021,Nature2021_2,Nature2022} is a distributed training paradigm that enables multiple clients to collaboratively train a global model without sharing their local data~\cite{federatedSurvy2023,federatedSurvy2024}. In a typical FL workflow, the server broadcasts a global model to clients at each communication round. Clients then update the model based on their local data and upload the updated parameters back to the server, which aggregates them to produce a new global model. As FL has been increasingly adopted in real-world applications, researchers have extended its applicability from homogeneous task settings to more general heterogeneous task scenarios. One important branch of this effort is \textbf{Federated Multi-Task Learning (FMTL)}~\cite{FMTLTPSD2021,FMTLAAAI2022,FMTLNIPS2017,FMTLNIPS2021}, which allows different clients to collaboratively train while performing inherently different tasks. 

\begin{figure}[t]
\centering
\includegraphics[width=\linewidth]{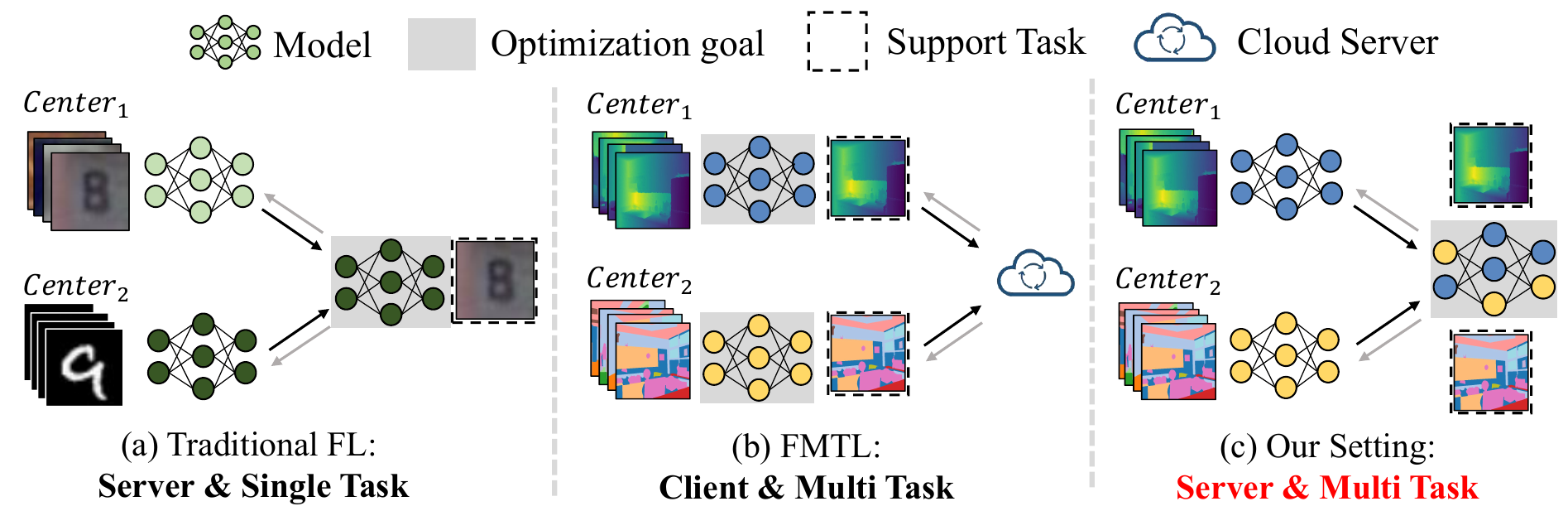}
\vspace{-10pt}
\caption{Conceptual comparison between traditional \textbf{FL}, \textbf{FMTL} and \textbf{our setting}. Unlike existing paradigms that either assume a shared task (a) or isolate multi-task training at the client level (b), our setting (c) enables task-disjoint clients to collaboratively train a unified global model through structure-aware aggregation on the server.}
\vspace{-15pt}
\label{fig:oursetting}
\end{figure}

Unlike conventional FL, as shown in Figure~\ref{fig:oursetting} (a), where all clients share the same task objective~\cite{noiidICCV2021,noiidCVPR2022,noiidICCV2023,noiidCVPR2023}. Existing FMTL methods are personalized federated learning methods that cannot achieve task aggregation among clients, as shown in Figure~\ref{fig:oursetting} (b). However, a more realistic demand is to fully leverage the heterogeneous data from all clients to \textbf{construct a unified global model} that supports all tasks, thereby reducing training costs, as shown in Figure~\ref{fig:oursetting} (c), called Heterogeneous Task Aggregation Federated Multi-Task Learning~(HTA-FMTL). In this scenario, not only do data distributions vary, but the tasks themselves may involve different objective functions, label structures, or even output spaces~\cite{MTLICCV2017,MTLICCV2023,MTLCVPR2023}. This heterogeneity poses a critical challenge: \textbf{Parameter Update Interference}. Since each client independently optimizes the model for its own task, the resulting update directions can differ significantly. Directly aggregating these updates may lead to interference of updates in parameter dimensions across tasks, thus diminishing the effectiveness of critical task-specific updates. This phenomenon severely hinders the optimization efficiency and multi-task adaptability of the global model.

To fully understand this phenomenon, we conduct a toy 
analysis experiment. Our experimental observations reveal that different tasks activate distinct regions of the model parameter space during training. This structural activation difference indicates that each task generates significant updates only in a subset of dimensions, while remaining silent in others. Based on this observation, we propose a key hypothesis: model updates generated by clients during local training can be decomposed into two components: \textbf{task-relevant updates} concentrated in task-specific subspaces, and \textbf{task-irrelevant perturbations} arising from redundant dimensions. The activation regions of task-relevant updates across different tasks are typically disjoint or even  approximately orthogonal~\cite{ilharco2022editing}, while task-irrelevant updates tend to fall within shared parameter regions, making them more susceptible to mutual interference during aggregation. This update structural inconsistency means that traditional aggregation strategies, which do not differentiate between update components, may mix critical task signals with irrelevant noise, leading to directional drift and the masking of important updates, ultimately degrading the performance of the global model on individual tasks. Furthermore, since all tasks are subject to similar interference, the model's performance across multiple tasks can be simultaneously impaired.

To address this challenge, we propose \textbf{FedDEA} (\textbf{F}ederated \textbf{DE}coupled \textbf{A}ggregation), a update-structure aware aggregation strategy designed to suppress task-irrelevant disturbances and enhance multi-task adaptability of the global model. At each communication round, FedDEA performs dimension filtering and rescaling, preserving task-responsive updates while amplifying their optimization effect, thereby achieving structural decoupling across tasks. Since FedDEA does not rely on task labels or modifications to model architectures, it can be easily integrated into mainstream federated optimization frameworks, offering strong generality and deployment-friendliness. 

Our main contributions are as follows:

\begin{itemize}
\item We systematically investigate and analyze a core challenge in HTA-FMTL: Parameter Update Interference, attribute its root cause to update structural disparities and mutual interference among task-support subspaces.

\item We propose FedDEA, a structure-aware aggregation strategy that integrates parameter decoupling and recalibration, effectively suppress update interference.

\item We conduct empirical studies on multiple task-heterogeneous datasets, and the results demonstrate that FedDEA, as a plug-in-play module, can be easily integrated into various federated optimization algorithms and consistently yields significant overall performance improvements.
\end{itemize}

\section{Related Work}
\label{sec:releated work}

\subsection{Federated Learning}
Federated learning is a representative distributed learning paradigm that enables multiple clients to collaboratively train a global model without sharing their local data. To address the widely observed issue of non-independent and identically distributed (non-IID) data, various strategies have been proposed. For example, FedProx~\cite{fedproxPMLS2020}, SCAFFOLD~\cite{ScaffoldPMLR2020}, and FedDyn~\cite{feddynICLR2021} introduce regularization terms in the local objective to mitigate client drift; FedNova~\cite{fednovaNIPS2020} applies normalization to balance training contributions across clients; FedOPT~\cite{fedoptICLR2021} employs adaptive optimizers to enhance convergence stability; MOON~\cite{moonCVPR2021} and FPLFPL~\cite{fplCVPR2023} adopt contrastive learning to improve local model consistency; and FedBN~\cite{fedbnICLR2021} aggregates only partial parameters to promote personalization. While these methods have shown effectiveness in alleviating data heterogeneity, they generally assume that all clients perform the same task~\cite{FMTLNIPS2017}. This assumption no longer holds in federated multi-task learning settings. When task objectives differ significantly, aggregating model updates indiscriminately can lead to gradient interference and parameter conflicts, ultimately degrading the global model's performance across tasks.

\subsection{Federated Multi-Task Learning}
Federated Multi-Task Learning (FMTL) aims to support collaborative training across clients that perform different tasks and has emerged as an important direction in personalized federated learning~\cite{PFMTLTPDS2021,PFMTLICML2021,PFMTLAAAI2021}. Early approaches such as MOCHA~\cite{FMTLNIPS2017}, FedEM~\cite{FMTLNIPS2021}, and FedSTA\cite{fedstaNips2021} primarily focus on personalized adaptation via local regularization or mixture-based modeling. More recent methods, including MaT-FL~\cite{MTFLCVPR2023}, FedBone~\cite{fedbonrJCST2024}, and FedHCA2~\cite{fedhca2CVPR2024}, further consider both task and data heterogeneity, proposing techniques such as shared representations, task partitioning, and cross-task aggregation to enhance collaboration across tasks. However, most existing methods still focus on optimizing personalized models locally or restrict aggregation to clients performing the same task~\cite{MTLSpringer1997,MTLSurvey2021TKDE}. They lack a unified modeling strategy that can support global collaboration across diverse tasks. This limitation becomes particularly evident in scenarios with non-overlapping task distributions or highly divergent update directions, where current approaches struggle to maintain coherent global performance.
e between tasks, thus driving better global optimization.

\section{Methodology}

\subsection{Problem Definition}

We consider \textbf{HTA-FMTL} setting, where the system consists of $K$ clients. Each client $k \in \{1, \dots, K\}$ holds a private local dataset $\mathcal{D}_k = \{(x_i^{(k)}, y_i^{(k)})\}$ and performs its own task $T_k$. Without sharing data, clients collaboratively train a unified model $w$ to optimize heterogeneous tasks.

The objective of FMTL is to minimize the task-aware global loss:
\begin{equation}
\min_{w} F(w) = \sum_{k=1}^K p_k F_k(w; T_k), \quad \text{where} \quad F_k(w; T_k) = \frac{1}{N_k} \sum_{i=1}^{N_k} \ell_{T_k}(x_i^{(k)}, y_i^{(k)}; w)
\end{equation}

Here, $\ell_{T_k}(\cdot)$ denotes the task-specific loss function corresponding to task $T_k$, and $p_k = \frac{N_k}{\sum_{j=1}^K N_j}$ is the weight proportional to the client’s data size.

\subsection{Motivation}

\paragraph{Observation of Parameter Update Interference.}

In HTA-FMTL, clients often perform {highly heterogeneous tasks}, which leads to {significant differences in local model updates}. These differences can interfere with the global aggregation process, a phenomenon we refer to as \textbf{Parameter Update Interference}. To systematically analyze this issue, we design a toy experiment involving four representative vision tasks: semantic segmentation, edge detection, surface normal estimation, and depth estimation, where each assigned to one of four clients for independent training. All clients use the same model architecture (Swin-T~\cite{SwinTransformerCVPR2021}), and the input data for each task comes from disjoint subsets of the overall dataset.

\begin{figure}[t]
\centering
\begin{subfigure}[t]{0.48\textwidth}
    \centering
    \includegraphics[width=\linewidth]{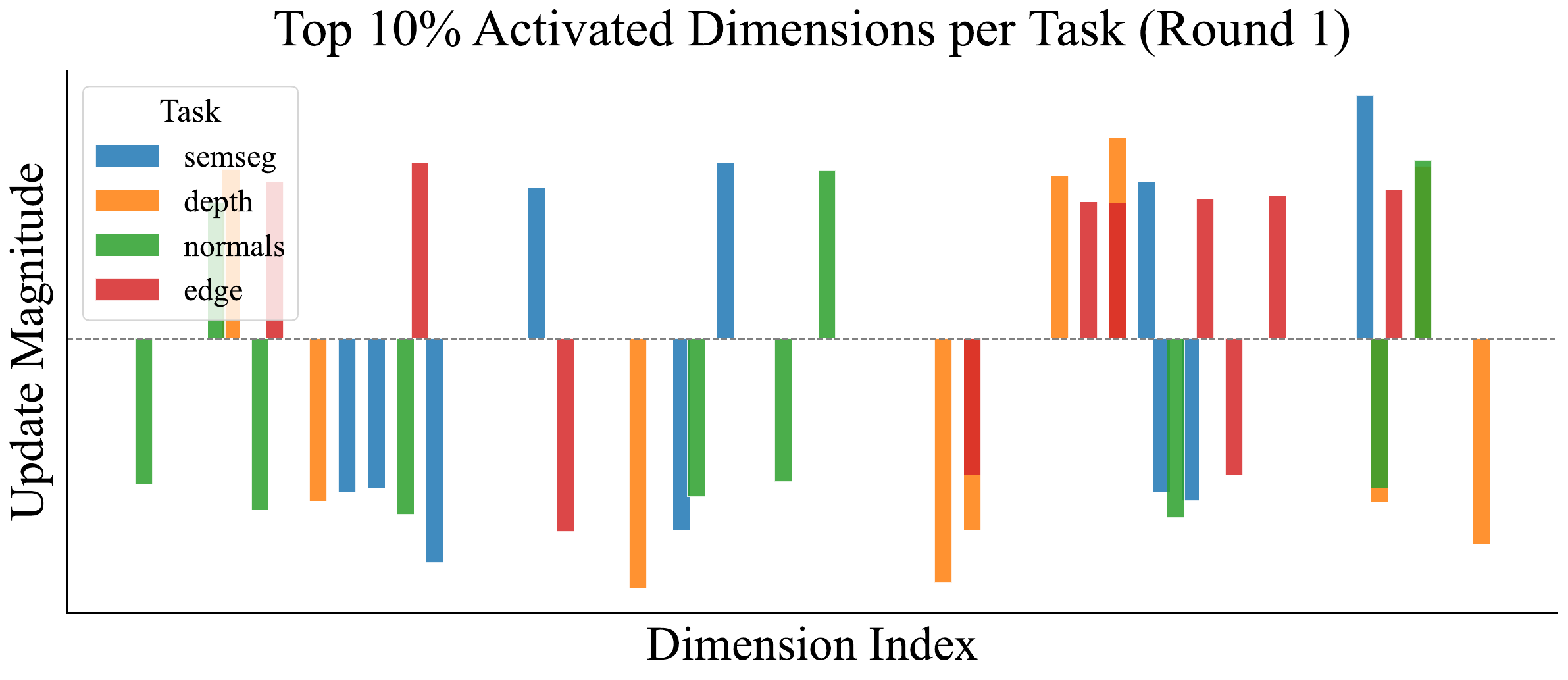}
    \caption{Round 1}
    \label{fig:mask-vis-a}
\end{subfigure}
\hfill
\begin{subfigure}[t]{0.48\textwidth}
    \centering
    \includegraphics[width=\linewidth]{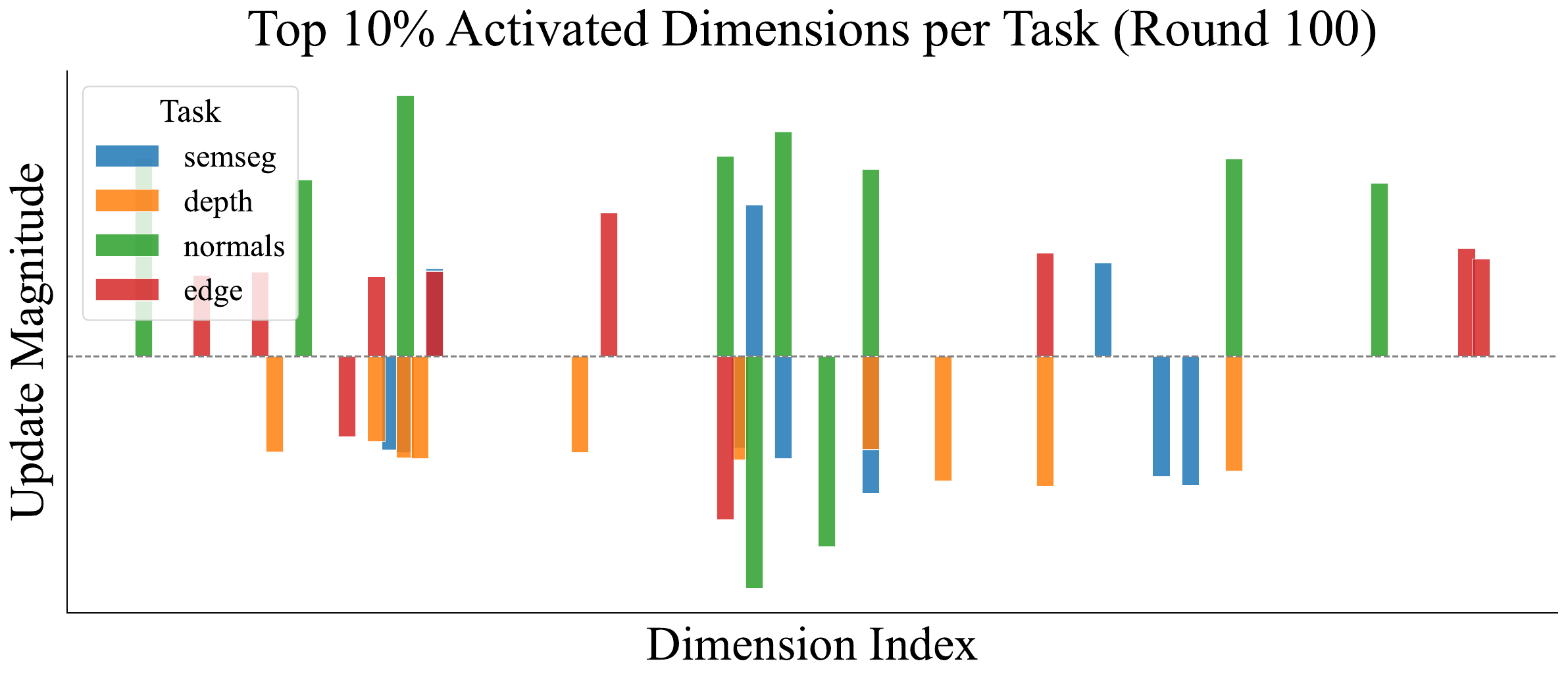}
    \caption{Round 100}
    \label{fig:mask-vis-b}
\end{subfigure}
\caption{Visualization of the top 10\% activated update dimensions per task in early and late communication rounds. \textbf{(a)} shows the parameter update distribution at Round 1, while \textbf{(b)} shows the distribution at Round 100.}
\label{fig:mask-vis}
\end{figure}

The experiment result, as shown in Figure~\ref{fig:mask-vis}, reveals that updates generated by different tasks are {concentrated in mutually exclusive subspaces}. These updates can be abstracted into two categories: (\emph{i}) {task-relevant signals}, which are confined to the task-support subspace and carry effective learning information, and (\emph{ii}) {task-irrelevant perturbations}, which reside in redundant dimensions. When such heterogeneous updates are aggregated without any filtering mechanism, the interference introduced in shared dimensions can lead to {directional drift}, weakening critical task signals. This ultimately results in {unstable training and degraded generalization performance} of the global model.
These observations provide clear insights for designing a better aggregation strategy: an ideal federated aggregation scheme should {explicitly filter out task-irrelevant components} to mitigate cross-task interference. However, {naive filtering inevitably compresses the update space}, which can reduce optimization energy and impair both convergence speed and final performance. Thus, a proper mechanism is also needed to {preserve the original update strength} in filtered dimensions. Based on the above analysis, we propose FedDEA (Federated Decoupled Aggregation), a structure-aware aggregation method designed to address update conflicts.
FedDEA first performs \textbf{parameter decoupling} by applying magnitude-based dimensional filtering, which approximates the task-specific support subspace by retaining only significant update dimensions while discarding redundant ones. To compensate for the potential reduction in update energy caused by this projection, FedDEA introduces a \textbf{recalibration mechanism} that uniformly rescales the retained updates, ensuring they maintain their original optimization strength in the projected subspace. This {two-stage mechanism of ``decoupling and recalibration''} constitutes the core of FedDEA, achieving better model aggregation.

\begin{figure}[t]
\centering
\includegraphics[width=\linewidth]{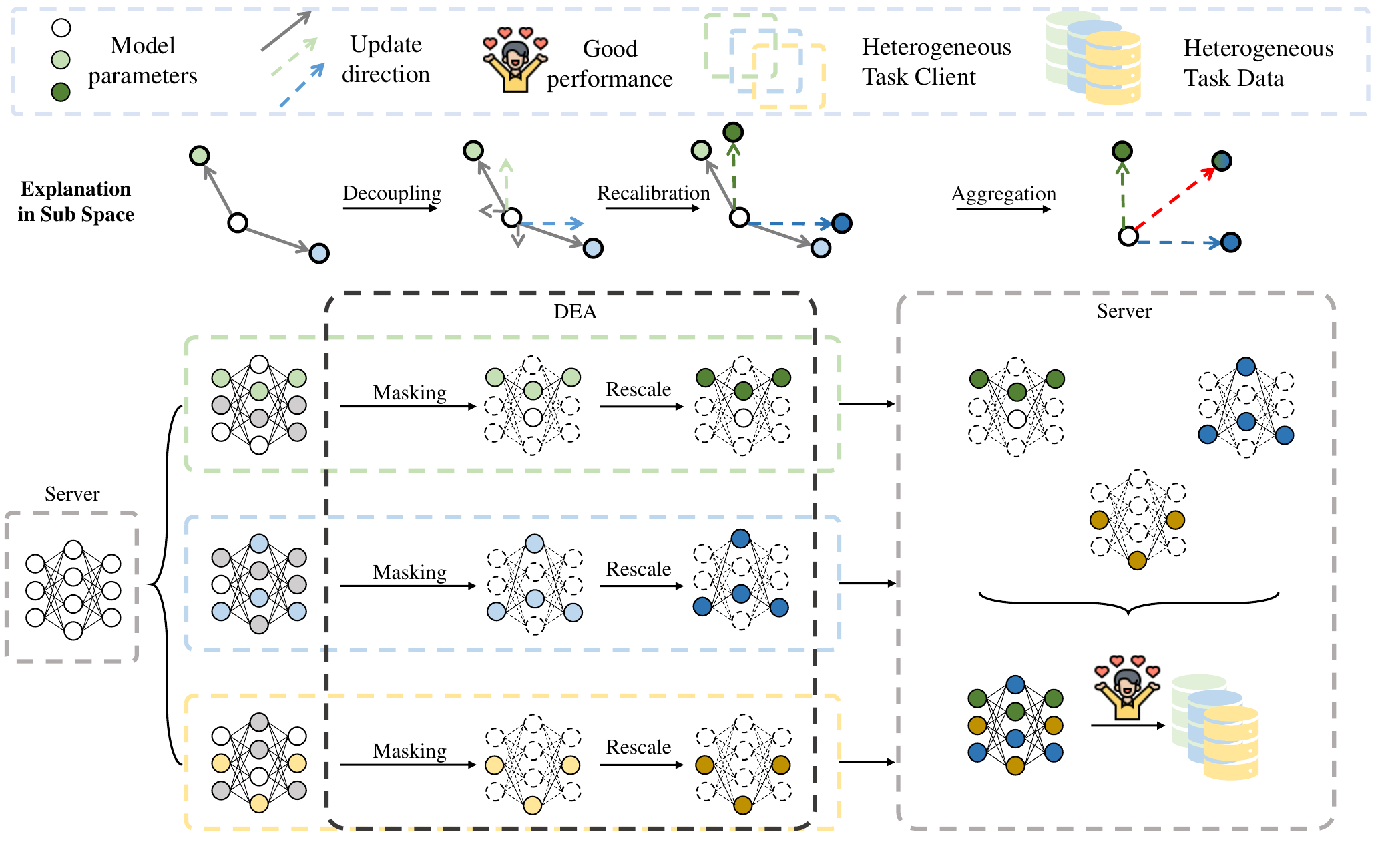}
\caption{Overview of the proposed method. Our approach addresses task heterogeneity in federated multi-task learning by structurally decoupling parameter updates across clients. Instead of directly aggregating all update dimensions, we introduce a structure-aware strategy that focuses on task-relevant subspaces. Specifically, each client performs local training and uploads its model updates; the server then applies a magnitude-based masking mechanism to retain only the most responsive dimensions for each task. These filtered updates are subsequently rescaled and aggregated, yielding a unified global model that preserves task-specific learning signals while suppressing cross-task interference.
}
\label{fig:method}
\end{figure}

\subsection{Proposed Method}

We propose {FedDEA}, a update structure-aware aggregation method tailored for HTA-FMTL. FedDEA performs explicit decoupling of task update paths through dimensional filtering and magnitude-based rescaling, effectively mitigating the problem of {parameter update interference}. By leveraging the differences in task-specific activation subspaces, FedDEA selectively aggregates only the significant dimensions that are highly relevant to each task, preserving effective optimization signals while suppressing redundant cross-task perturbations. Framework of our method  is shown in Figure~\ref{fig:method}, which consists of two core components: parameter decoupling and recalibration.

\paragraph{Parameter Decoupling.} In a federated learning system with $K$ clients, the global model at round $t$ is denoted as $\boldsymbol{\theta}^{(t)} \in \mathbb{R}^d$. Each clien $k$ erforms local training using its private dataset $\mathcal{D}_k$ resulting in an updated local model $\boldsymbol{\theta}_k^{(t)}$. The corresponding local update vector is then computed as:
\begin{equation}
    \boldsymbol{\Delta}_k^{(t)} = \boldsymbol{\theta}_k^{(t)} - \boldsymbol{\theta}^{(t)},
\end{equation}
Let  $\boldsymbol{\Delta}_k^{(t)} \in \mathbb{R}^d$ denote the model update of client $k$ in round $t$.

To extract task-relevant update signals, we introduce a hyperparameter $\rho \in (0, 1]$(selection ratio), referred to as the selection ratio, which controls the proportion of update dimensions retained by each client. Specifically, among all elements in $\boldsymbol{\Delta}_k^{(t)}$, we identify the top $\lfloor \rho d \rfloor$ dimensions with the largest absolute magnitudes and retain them, while setting the remaining positions to zero. Let $\mathcal{I}_k^{\rho}$ denote the index set of these selected dimensions. We then construct a binary mask vector  $\mathbf{m}_k \in \{0,1\}^d$ as follows:
\begin{equation}
    m_k^{(i)} =
    \begin{cases}
    1, & i \in \mathcal{I}_k^{\rho}, \\
    0, & \text{otherwise}.
    \end{cases}
\end{equation}

The update obtained through the masking operation is given by:
\begin{equation}
\widetilde{\boldsymbol{\Delta}}_k^{(t)} = \mathbf{m}_k \odot \boldsymbol{\Delta}_k^{(t)},
\end{equation}
where $\odot$ denotes the element-wise multiplication. This process effectively projects the original update $\boldsymbol{\Delta}_k^{(t)}$ onto the support subspace of its local task, thereby filtering out noise from low-response dimensions and reducing the interference propagation across inconsistent task dimensions.

\paragraph{Recalibration.} However, since the dimensionality selection compresses the effective update region, the overall magnitude of the selected update becomes smaller, potentially weakening the driving force of global training. To address this issue, we apply a uniform rescaling to the retained dimensions in order to restore their representational strength, as follows:
\begin{equation}
    \widetilde{\boldsymbol{\Delta}}_k^{(t)} \leftarrow \frac{1}{\rho} \cdot \widetilde{\boldsymbol{\Delta}}_k^{(t)}.
\end{equation}
This rescaling strategy approximately preserves the original energy level of the update {without altering its direction}, thereby ensuring dynamic consistency during the global aggregation process.

Finally, all clients upload their processed update $\widetilde{\boldsymbol{\Delta}}_k^{(t)}$ to the server. The server performs global model updating using the conventional weighted aggregation rule:
\begin{equation}
    \boldsymbol{\theta}^{(t+1)} = \boldsymbol{\theta}^{(t)} + \cdot \sum_{k=1}^{K} w_k \cdot \widetilde{\boldsymbol{\Delta}}_k^{(t)},
\end{equation}
where $w_k$ denotes the aggregation weight for client $k$, typically defined as $w_k = \frac{|\mathcal{D}_k|}{\sum_{j=1}^K |\mathcal{D}_j|}$, i.e., normalized by the local data size.

\section{Experiment}

This section aims to address the following two research questions through empirical investigation:

\begin{itemize}
\item \textbf{RQ1:} Can FedDEA construct \textbf{a unified global model} that can perform each client's heterogeneous task? Is this capability beyond what existing personalization or multi-task federated learning methods can achieve?
\item \textbf{RQ2:} Is FedDEA generalizable enough to serve as \textbf{a plug-in-play module} that can be integrated into various federated optimization strategies? What are the advantages of the DEA mechanism compared to traditional and existing masking-based methods?
\end{itemize}

\subsection{Experimental Setup}

\paragraph{Dataset.} Following the settings in prior work  ~\cite{fedhca2CVPR2024,MTFLCVPR2023}, we evaluate FedDEA on two standard multi-task vision datasets: \textbf{NYUD-V2}~\cite{NYUD-V2} and \textbf{PASCAL-CONTEXT}~\cite{PASCAL-context}. NYUD-V2 includes four tasks: semantic segmentation(Semseg), depth estimation(Depth), normal detection(Normals) and edge detection(Edge); PASCAL-CONTEXT contains five tasks: semantic segmentation(Semseg), human-parts segmentation(Parts), normal detection(Normals), saliency estimation(Sal) and edge detection(Edge). For more detailed information, please refer supplementary material.

\paragraph{Model Architecture \& Federated Settings.} In the main experiments, we use a multi-decoder model with \textbf{Swin-T}~\cite{SwinTransformerCVPR2021} as the encoder and a simple FCN~\cite{FCNCVPR2015} as the decoder. Additionally, in the supplementary material, we include extended experiments using \textbf{ResNet-18}~\cite{he2016deep} as the backbone to analyze the robustness of our method across different model architectures. All clients share the same model architecture and follow a unified training protocol for federated optimization. Detailed settings are provided in supplementary material.

\paragraph{Evaluation Metrics.}  Following ~\cite{MTLSpringer1997,MTLSurvey2021TKDE,MTLNature2024,MTLSurvey2024ACM,MTLSurveyCoRR2024}, we use mIoU for semantic segmentation, RMSE for depth estimation, mErr for normal detection, Sal for saliency estimation, and odsF for edge detection.

\subsection{Unified Global Model Construction Capability (RQ1)}

\begin{table}[ht]\small
\centering
\begin{minipage}[t]{0.48\textwidth}
\caption{Comparison with representative federated personalization and multi-task methods on NYUD-V2. “\textbackslash” ~indicates that the corresponding model is unable to perform the given task. Unlike others that require separate models per task, our method supports multiple tasks within a single unified model, enabling efficient and scalable deployment.}
\renewcommand\arraystretch{1.1}
\begin{adjustbox}{max width=\linewidth}
\begin{tabular}{c||cccc}
\hline\thickhline
\rowcolor{lightgray}
& \multicolumn{4}{c}{\textbf{NYUD}} \\
\cline{2-5}
\rowcolor{lightgray}
\multirow{-2}{*}{\textbf{Method}} 
& \makecell{Semseg \\ (mIoU)$\uparrow$} 
& \makecell{Depth \\ (RMSE)$\downarrow$} 
& \makecell{Normals \\ (mErr)$\downarrow$} 
& \makecell{Edge \\ (odsF)$\uparrow$} \\
\arrayrulecolor{gray!60}\Xhline{0.8pt}
\arrayrulecolor{black}
Local(Semseg)    & 31.15 & \textbackslash & \textbackslash & \textbackslash \\
Local(Depth)     & \textbackslash & 0.7087 & \textbackslash & \textbackslash \\
Local(Normals)   & \textbackslash & \textbackslash & 24.59 & \textbackslash \\
Local(Edge)      & \textbackslash & \textbackslash & \textbackslash & 73.86 \\
\hline\hline
FedAMP(Semseg)~\cite{fedampAAAI2021}   & 33.86 & \textbackslash & \textbackslash & \textbackslash \\
FedAMP(Depth)~\cite{fedampAAAI2021}    & \textbackslash & 0.7500 & \textbackslash & \textbackslash \\
FedAMP(Normals)~\cite{fedampAAAI2021}  & \textbackslash & \textbackslash & 23.64 & \textbackslash \\
FedAMP(Edge)~\cite{fedampAAAI2021}     & \textbackslash & \textbackslash & \textbackslash & 75.19 \\
\hline\hline
FedMTL(Semseg)~\cite{FMTLNIPS2017}   & 32.47 & \textbackslash & \textbackslash & \textbackslash \\
FedMTL(Depth)~\cite{FMTLNIPS2017}    & \textbackslash & 0.7674 & \textbackslash & \textbackslash \\
FedMTL(Normals)~\cite{FMTLNIPS2017}  & \textbackslash & \textbackslash & 23.58 & \textbackslash \\
FedMTL(Edge)~\cite{FMTLNIPS2017}     & \textbackslash & \textbackslash & \textbackslash & 75.28 \\
\hline
FedHCA2(Semseg)~\cite{fedhca2CVPR2024}  & 32.93 & \textbackslash & \textbackslash & \textbackslash \\
FedHCA2(Depth)~\cite{fedhca2CVPR2024}   & \textbackslash & 0.7227 & \textbackslash & \textbackslash \\
FedHCA2(Normals)~\cite{fedhca2CVPR2024} & \textbackslash & \textbackslash & 23.67 & \textbackslash \\
FedHCA2(Edge)~\cite{fedhca2CVPR2024}    & \textbackslash & \textbackslash & \textbackslash & 75.17 \\
\hline\hline
\textbf{FedAVG+DEA(Ours)} & 30.78 & 0.7052 & 24.62 & 74.77 \\
\textbf{FedProx+DEA(Ours)} & 31.46 & 0.6958 & 24.42 & 74.83 \\
\thickhline
\end{tabular}
\end{adjustbox}
\label{tab:rq1-main-table-nyud}
\end{minipage}
\hfill
\begin{minipage}[t]{0.494\textwidth}
\caption{Comparison with representative feder-
ated personalization and multi-task methods on
PASCAL-CONTEXT. “\textbackslash”~ indicates that the corresponding model is unable to perform the given task. Unlike others that require separate models per task, our method supports
multiple tasks within a single unified model.}
\renewcommand\arraystretch{1.1}
\begin{adjustbox}{max width=\linewidth}
\begin{tabular}{c||ccccc}
\hline\thickhline
\rowcolor{lightgray}
& \multicolumn{5}{c}{\textbf{PASCAL}} \\
\cline{2-6}
\rowcolor{lightgray}
\multirow{-2}{*}{\textbf{Method}} 
& \makecell{Semseg \\ (mIoU)$\uparrow$} 
& \makecell{Parts \\ (mIoU)$\uparrow$} 
& \makecell{Normals \\ (mErr)$\downarrow$} 
& \makecell{Sal \\ (maxF)$\uparrow$} 
& \makecell{Edge \\ (odsF)$\uparrow$} \\
\arrayrulecolor{gray!60}\Xhline{0.8pt}
\arrayrulecolor{black}
Local(Semseg)      & 50.48 & \textbackslash & \textbackslash & \textbackslash & \textbackslash \\
Local(Parts)       & \textbackslash & 49.86 & \textbackslash & \textbackslash & \textbackslash \\
Local(Normals)     & \textbackslash & \textbackslash & 16.23 & \textbackslash & \textbackslash \\
Local(maxF)        & \textbackslash & \textbackslash & \textbackslash & 78.20 & \textbackslash \\
Local(Edge)        & \textbackslash & \textbackslash & \textbackslash & \textbackslash & 68.20 \\
\hline\hline
FedAMP(Semseg)~\cite{fedampAAAI2021}     & 60.05 & \textbackslash & \textbackslash & \textbackslash & \textbackslash \\
FedAMP(Parts)~\cite{fedampAAAI2021}      & \textbackslash & 51.93 & \textbackslash & \textbackslash & \textbackslash \\
FedAMP(Normals)~\cite{fedampAAAI2021}    & \textbackslash & \textbackslash & 16.23 & \textbackslash & \textbackslash \\
FedAMP(maxF)~\cite{fedampAAAI2021}       & \textbackslash & \textbackslash & \textbackslash & 79.49 & \textbackslash \\
FedAMP(Edge)~\cite{fedampAAAI2021}       & \textbackslash & \textbackslash & \textbackslash & \textbackslash & 70.21 \\
\hline\hline
FedMTL(Semseg)~\cite{FMTLNIPS2017}     & 59.89 & \textbackslash & \textbackslash & \textbackslash & \textbackslash \\
FedMTL(Parts)~\cite{FMTLNIPS2017}      & \textbackslash & 51.35 & \textbackslash & \textbackslash & \textbackslash \\
FedMTL(Normals)~\cite{FMTLNIPS2017}    & \textbackslash & \textbackslash & 15.77 & \textbackslash & \textbackslash \\
FedMTL(maxF)~\cite{FMTLNIPS2017}       & \textbackslash & \textbackslash & \textbackslash & 79.55 & \textbackslash \\
FedMTL(Edge)~\cite{FMTLNIPS2017}       & \textbackslash & \textbackslash & \textbackslash & \textbackslash & 69.85 \\
\hline
FedHCA2(Semseg)~\cite{fedhca2CVPR2024}    & 58.25 & \textbackslash & \textbackslash & \textbackslash & \textbackslash \\
FedHCA2(Parts)~\cite{fedhca2CVPR2024}     & \textbackslash & 52.33 & \textbackslash & \textbackslash & \textbackslash \\
FedHCA2(Normals)~\cite{fedhca2CVPR2024}   & \textbackslash & \textbackslash & 15.73 & \textbackslash & \textbackslash \\
FedHCA2(maxF)~\cite{fedhca2CVPR2024}      & \textbackslash & \textbackslash & \textbackslash & 79.43 & \textbackslash \\
FedHCA2(Edge)~\cite{fedhca2CVPR2024}      & \textbackslash & \textbackslash & \textbackslash & \textbackslash & 70.25 \\
\hline\hline
\textbf{FedAVG+DEA(Ours)}  & 55.23 & 51.18 & 16.69 & 79.24 & 69.25 \\
\textbf{FedProx+DEA(Ours)} & 54.73 & 51.40 & 16.64 & 79.73 & 69.42 \\
\thickhline
\end{tabular}
\end{adjustbox}
\label{tab:rq1-main-table-pas}
\end{minipage}
\end{table}

This experiment aims to verify whether FedDEA can construct \textbf{a unified global model} capable of adapting to multiple tasks in HTA-FMTL. We compare three representative methods: \textbf{FedAMP}~(Personlized Method)~\cite{fedampAAAI2021}, \textbf{FedMTL}~(Multi-task Method)~\cite{FMTLNIPS2017} and \textbf{FedHCA2}~(Multi-task Method)~\cite{fedhca2CVPR2024}. Notably, these three methods generate independent models for each task and thus cannot produce a unified global parameter representation. According to Table~\ref{tab:rq1-main-table-nyud} and Table~\ref{tab:rq1-main-table-pas}, we observe the following:

\begin{itemize}
\item \textbf{Obs1:} Although certain personalized and multi-task federated methods achieve slightly better performance than FedDEA on some task-specific metrics, these methods optimize each task individually and \textbf{cannot produce a unified model that can be shared across multiple tasks}. In contrast, FedDEA outputs a global model that \textbf{structurally adapts to multiple tasks within a single model}.
\end{itemize}

\begin{table*}[!t]
\caption{Performance comparison across NYUD and PASCAL datasets using different enhancement modules. We compare our method with two techniques, PCGrad~\cite{pcgrad} and FedHEAL~\cite{fedhealCVPR2024}. $\Delta$ represents the overall performance gain brought by adding the proposed module, compared to the original method. "\--{}" indicates the baseline result without any enhancement, so no gain is computed. While PCGrad~\cite{pcgrad} and FedHEAL~\cite{fedhealCVPR2024} often lead to performance degradation or negligible improvement for most baseline methods, our approach consistently improves all baselines, highlighting its superior generalization and cross-task compatibility.}
\small
\setlength{\abovecaptionskip}{0cm}
\centering
\resizebox{\linewidth}{!}{
\renewcommand\arraystretch{1.1}
\begin{tabular}{r||cccc|c||ccccc|c}
\hline\thickhline
\rowcolor{lightgray}
& \multicolumn{5}{c||}{\textbf{NYUD}} & \multicolumn{6}{c}{\textbf{PASCAL}} \\
\cline{2-12}
\rowcolor{lightgray}
\multirow{-2}{*}{\textbf{Method}} 
& \makecell{Semseg \\ (mIoU)$\uparrow$} 
& \makecell{Depth \\ (RMSE)$\downarrow$} 
& \makecell{Normals \\ (mErr)$\downarrow$} 
& \makecell{Edge \\ (odsF)$\uparrow$} 
& $\Delta \%$ 
& \makecell{Semseg \\ (mIoU)$\uparrow$} 
& \makecell{Parts \\ (mIoU)$\uparrow$} 
& \makecell{Normals \\ (mErr)$\downarrow$} 
& \makecell{Sal \\ (maxF)$\uparrow$} 
& \makecell{Edge \\ (osdF)$\uparrow$} 
& $\Delta \% $ \\
\arrayrulecolor{gray!60}\Xhline{0.8pt}
\arrayrulecolor{black}

FedAvg~\cite{fedavg2017}        & 23.05 & 0.7213 & 26.52 & 75.19 & -- & 47.22 & 44.55 & 17.45 & 79.09 & 69.66 & -- \\
+PCGrad~\cite{pcgrad}       & 16.48 & 1.5323 & 29.64 & 74.59 & -38.37 & 26.01 & 41.53 & 22.35 & 77.94 & 68.85 & -16.48 \\
+HEAL~\cite{fedhealCVPR2024}         & 23.37 & 0.7628 & 26.65 & 74.50 & -1.44  & 46.22 & 37.46 & 17.30 & 78.69 & 69.47 & -3.59 \\
\textbf{+DEA(Ours)   }       & 30.78 & 0.7052 & 24.62 & 74.77 &\textbf{+10.60}  & 55.23 & 51.18 & 16.69 & 79.24 & 69.25 &\textbf{+ 7.16} \\
\hline
FedProx~\cite{fedproxPMLS2020}       & 22.57 & 0.7104 & 26.97 & 75.16 & --       & 38.40 & 43.79 & 17.49 & 79.76 & 69.58 & -- \\
+PCGrad~\cite{pcgrad}       & 16.45 & 1.5227 & 29.05 & 74.16 & -37.62 & 29.41 & 39.45 & 21.44 & 77.98 & 69.26 & -11.72 \\
+HEAL~\cite{fedhealCVPR2024}         & 21.57 & 0.7545 & 25.48 & 74.25 & -1.58  & 36.95 & 42.68 & 17.37 & 79.44 & 69.40 & -1.26 \\
\textbf{+DEA(Ours) }         & 32.36 & 0.7229 & 24.61 & 75.31 &\textbf{+12.65}  & 54.73 & 51.40 & 16.64 & 79.73 & 69.42 & \textbf{+12.89} \\
\hline
FedNova~\cite{fednovaNIPS2020}       & 24.78 & 0.7195 & 26.35 & 75.20 & --       & 41.83 & 46.77 & 17.67 & 78.77 & 69.67 & -- \\
+PCGrad~\cite{pcgrad}       & 5.56  & 0.7411 & 30.20 & 74.29 & -24.10 & 7.29  & 29.66 & 20.94 & 78.13 & 69.69 & -27.68 \\
+HEAL~\cite{fedhealCVPR2024}         & 23.16 & 0.7044 & 26.58 & 74.28 & -1.63  & 47.49 & 48.14 & 17.39 & 78.76 & 68.70 & +3.33 \\
\textbf{+DEA(Ours)}          & 30.37 & 0.7160 & 24.75 & 74.50 & \textbf{+7.05}   & 49.01 & 46.46 & 17.09 & 79.08 & 69.72 & \textbf{+4.05} \\
\hline
FedDyn~\cite{feddynICLR2021}        & 22.81 & 0.7234 & 26.68 & 75.28 & --       & 45.21 & 43.34 & 17.53 & 78.90 & 69.76 & -- \\
+PCGrad~\cite{pcgrad}       & 14.12 & 1.5366 & 29.48 & 74.29 & -40.56 & 23.36 & 39.12 & 21.87 & 77.14 & 68.40 & -17.41 \\
+HEAL~\cite{fedhealCVPR2024}         & 21.92 & 0.7119 & 26.84 & 74.25 & -1.06  & 44.55 & 40.42 & 17.52 & 78.72 & 69.59 & -1.73 \\
\textbf{+DEA(Ours)  }        & 31.11 & 0.7232 & 24.66 & +74.50 & \textbf{+10.75}  & 56.20 & 50.35 & 16.50 & 79.27 & 69.01 & \textbf{+9.15} \\
\hline
Moon~\cite{moonCVPR2021}          & 13.66 & 0.7192 & 26.06 & 75.19 & --       & 22.18 & 36.90 & 17.98 & 78.83 & 68.40 & -- \\
+PCGrad~\cite{pcgrad}       & 18.85 & 1.7297 & 30.31 & 74.10 & -30.06 & 40.84 & 40.01 & 17.53 & 79.51 & 68.72 & -2.80 \\
+HEAL~\cite{fedhealCVPR2024}         & 13.70 & 0.7018 & 26.49 & 74.26 & -0.04  & 42.18 & 41.90 & 17.27 & 78.83 & 68.40 & -1.30 \\
\textbf{+DEA(Ours) }         & 26.15 & 0.7738 & 25.66 & 74.25 &\textbf{ 21.03}  & 46.76 & 45.22 & 16.99 & 79.57 & 68.43 & \textbf{+2.84} \\
\hline
FedBN~\cite{fedbnICLR2021}         & 24.42 & 0.7174 & 26.90 & 75.06 & --       & 43.18 & 44.04 & 17.69 & 78.55 & 69.84 & -- \\
+PCGrad~\cite{pcgrad}       & 19.26 & 1.4769 & 29.64 & 74.26 & -34.56 & 42.15 & 44.22 & 17.66 & 78.54 & 69.59 & -0.44 \\
+HEAL~\cite{fedhealCVPR2024}         & 23.57 & 0.7158 & 26.65 & 74.65 & -0.71  & 42.48 & 41.02 & 17.37 & 78.99 & 69.56 & -1.30 \\
\textbf{+DEA(Ours) }         & 30.32 & 0.7179 & 24.53 & 74.68 & \textbf{+8.10}   & 52.48 & 47.03 & 17.26 & 78.45 & 69.82 & \textbf{+6.12} \\
\hline
FedGA~\cite{fedgaCVPR2023}         & 26.49 & 0.7419 & 55.30 & 75.46 & --       & 53.80 & 43.49 & 87.75 & 78.47 & 70.46 & -- \\
+PCGrad~\cite{pcgrad}       & 4.71  & 1.8604 & 80.22 & 74.29 & -69.90 & 8.91 	&20.27 	&93.03 	&43.09 	& 68.56 
     & -38.13 \\
+HEAL~\cite{fedhealCVPR2024}         & 22.87 & 0.7274 & 48.20 & 75.16 & +0.18   &44.55 	&40.42 	&88.31 	&78.72 	&69.76 
 
     & -5.12 \\
\textbf{+DEA(Ours) }         & 28.07 & 0.7331 & 55.91 & 75.44 & \textbf{+1.50}   & 56.38 & 45.46 & 86.03 & 78.56 & 70.04 & \textbf{+2.16} \\
\hline
FedACG~\cite{fedacgCVPR2024}        & 31.54 & 0.7526 & 25.76 & 75.19 & --       & 54.61 & 50.33 & 17.27 & 78.82 & 70.42 & -- \\
+PCGrad~\cite{pcgrad}       & 31.35 & 0.7134 & 25.46 & 74.25 & +1.13   & 54.73 & 50.12 & 16.79 & 79.85 & 70.02 & +0.67 \\
+HEAL~\cite{fedhealCVPR2024}           & 31.84 & 0.7057 & 25.78 & 74.28 & +1.47   & 55.65 & 50.01 & 17.11 & 78.68 & 69.83 & +0.24 \\
\textbf{+DEA(Ours)  }        & 32.51 & 0.7234 & 25.19 & 74.93 & \textbf{+2.20}   & 56.86 & 51.39 & 16.94 & 78.95 & 70.04 & \textbf{+1.55} \\
\hline

\thickhline
\end{tabular}
}
\label{tab:rq2-comparison}
\end{table*}

\subsection{Generalizability Across Different Aggregation Strategies (RQ2)}

This experiment aims to verify whether FedDEA can serve as \textbf{a universal plug-in-play module} that can be easily integrated into mainstream aggregation strategies without modifying the core optimization process. To this end, we select eight representative federated optimization methods as base algorithms: FedAvg~\cite{fedavg2017}, FedProx~\cite{fedproxPMLS2020}, FedNova~\cite{fednovaNIPS2020}, MOON~\cite{moonCVPR2021}, FedDyn~\cite{feddynICLR2021}, FedBN~\cite{fedbnICLR2021}, FedGA, and FedACG~\cite{fedacgCVPR2024}, and integrate FedDEA into each of them. In addition, we include both traditional masking methods and existing masking methods as baseline comparisons: PCGrad~\cite{pcgrad} and FedHEAL~\cite{fedhealCVPR2024}.

In HTA-FMTL, the evaluation criteria vary across tasks, making it difficult to perform unified comparisons using traditional metrics. Following ~\cite{deltam}, we adopt $\Delta$ to measure the overall performance gain brought by the enhancement module across tasks.

\begin{equation}
\Delta = \frac{1}{T} \sum_{t=1}^{T} \alpha_t \cdot \frac{S_t^{\text{+}} - S_t^{\text{base}}}{S_t^{\text{base}}} \times 100 \%
\end{equation}

Here, $T$ denotes the number of task metrics, $S_t^{\text{base}}$ and $S_t^{\text{+}}$ represent the performance scores before and after integrating the enhancement module, respectively. If the $t$-th metric follows a "higher is better" criterion, then $\alpha_t = +1$; if it follows a "lower is better" criterion, then $\alpha_t = -1$. 

The experimental results are presented in Table~\ref{tab:rq2-comparison}, from which we make the following observations:
\begin{itemize}
    \item \textbf{Obs2:} Across all  federated optimization methods, integrating FedDEA \textbf{consistently brings stable and significant performance improvements}, with $\Delta$ remaining positive throughout. This demonstrates that FedDEA, as a universal enhancement module, possesses strong adaptability and consistency. Moreover, FedDEA can be easily embedded into existing aggregation workflows without modifying model architectures or introducing additional communication overhead, resulting in low deployment cost and high engineering feasibility.
    
    \item \textbf{Obs3:} Compared to PCGrad~\cite{pcgrad},\textbf{ FedDEA offers stronger structural modeling for task heterogeneity}. PCGrad mitigates directional conflicts via projection but ignores dimension-level differences in the parameter space. When tasks activate non-overlapping subspaces, it may retain irrelevant updates, leading to interference or degradation. FedDEA instead performs decoupling directly in the parameter subspace, effectively avoiding such conflicts at their source.
    
    \item \textbf{Obs4:} Compared to FedHEAL~\cite{fedhealCVPR2024}, \textbf{FedDEA shows greater adaptability}. FedHEAL depends on server-side consistency checks using historical updates, which can suppress critical task-specific gradients in heterogeneous settings, causing performance drops. In contrast, FedDEA uses local update significance to generate masks without centralized coordination or label reliance, enabling more accurate task-specific updates and more robust multi-task enhancement.

\end{itemize}

\subsection{Methodological Analysis Experiments}

To gain a more comprehensive understanding of the specific contributions of FedDEA’s key design components to model performance, we conducted a series of methodological analysis experiments. These evaluations focus on ablation study, hyperparameter sensitivity and convergence analysis.

\begin{wraptable}{r}{0.55\textwidth} 
\vspace{-1ex}  
\centering
\caption{Ablation study of DEA on NYUD-V2 dataset.}
\label{tab:glad-ablation-nyud}
\renewcommand\arraystretch{1.1}
\begin{adjustbox}{max width=0.53\textwidth}
\begin{tabular}{l||cccc|c}
\hline\thickhline
\rowcolor{lightgray}
& \multicolumn{5}{c}{\textbf{NYUD}} \\
\cline{2-6}
\rowcolor{lightgray}
\multirow{-2}{*}{\textbf{Method}} 
& \makecell{Semseg \\ (mIoU)$\uparrow$} 
& \makecell{Depth \\ (RMSE)$\downarrow$} 
& \makecell{Normals \\ (mErr)$\downarrow$} 
& \makecell{Edge \\ (odsF)$\uparrow$} 
& $\Delta$m\% \\
\arrayrulecolor{gray!60}\Xhline{0.8pt}
\arrayrulecolor{black}
FedAvg     & 23.05  & 0.7213  & 26.52  & \textbf{75.19} & -         \\
+DEA(a)   & 1.22   & 2.1058  & 52.40  & 57.44          & -101.96   \\
+DEA(b)   & 11.49  & 0.7351  & 27.74  & 75.17          & -14.17    \\
+DEA(c)   & 24.41  & 0.7352  & 26.90  & 75.18          & 0.64      \\
+DEA      & \textbf{30.78} & \textbf{0.7052} & \textbf{24.62} & 74.77 & \textbf{10.60} \\
\hline\hline
\end{tabular}
\end{adjustbox}
\vspace{-1ex} 
\end{wraptable}

\paragraph{Ablation Analysis of Decoupling and Recalibration Mechanisms.}
To systematically evaluate the effectiveness of FedDEA's core mechanisms, we designed three variants to analyze the impact of its masking and rescaling strategies. \textbf{(a) Small Mask} retains only the dimensions with the smallest gradient magnitudes; \textbf{(b) No Rescale} skips the rescaling step; \textbf{(c) Random Mask} randomly selects parameter dimensions. Experimental results in Table~\ref{tab:glad-ablation-nyud} show that update structure-unaware masking significantly degrades performance, while the absence of rescaling leads to slower convergence and reduced final accuracy.

\begin{figure}[htbp]
  \centering
  \begin{minipage}[t]{0.56\textwidth}
    \vspace{0pt} 
    \paragraph{Effect of Selection Rate $\rho$.} 
    To assess the impact of the selection rate $\rho$, we conducted experiments on the NYUD-V2 dataset using FedAvg and FedProx as baselines. We varied $\rho$ from 10\% to 100\% and recorded the performance trends of each task metric, along with changes in the overall improvement $\Delta$. As shown in Figure~\ref{fig:selection_rate}, effect of $\rho$ on $\Delta$ follows a general increasing then decreasing pattern, confirming that structural filtering under a reasonable compression ratio helps concentrate updates on critical dimensions, improves aggregation quality, and serves as a key prerequisite for effective multi-task modeling.
  \end{minipage}%
  \hfill
  \begin{minipage}[t]{0.38\textwidth}
    \vspace{0pt}  
    \includegraphics[width=\linewidth]{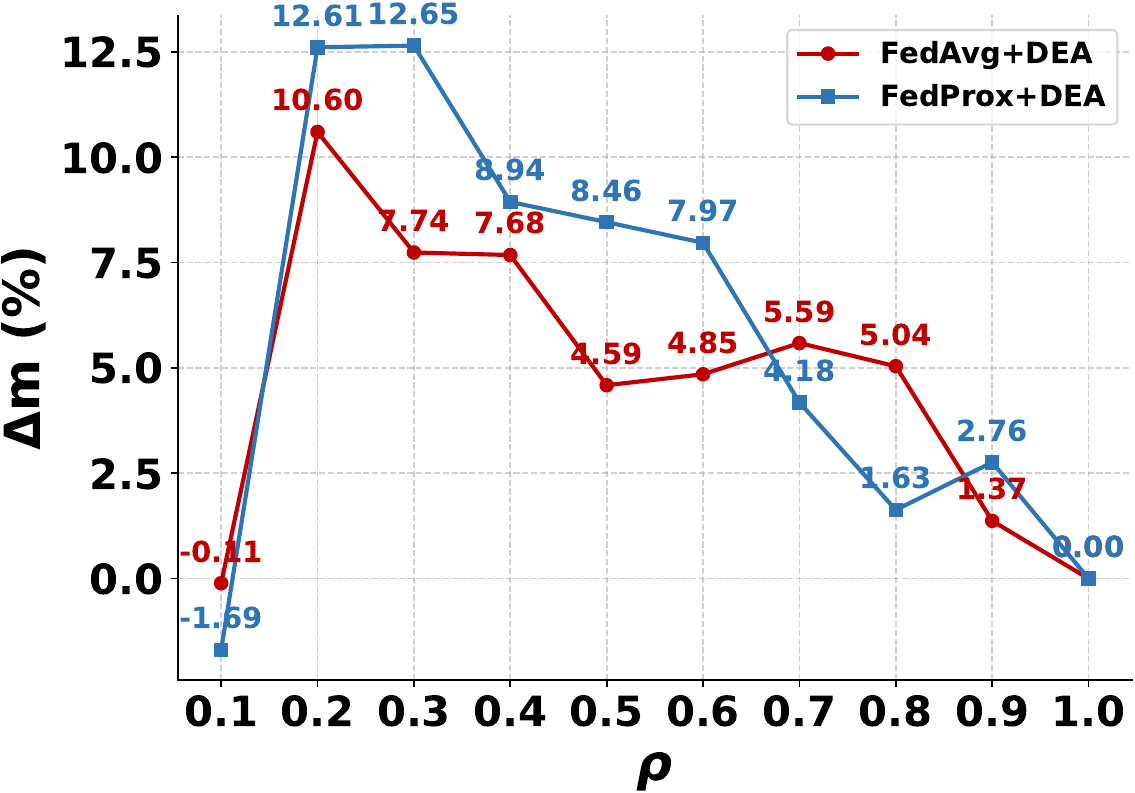}
    \vspace{-15pt}
    \caption{Hyperparameter analysis of selection rate $\rho$.}
    \vspace{-15pt}
    \label{fig:selection_rate}
  \end{minipage}
\end{figure}

\begin{figure}[ht]
\centering
\begin{minipage}[t]{0.48\textwidth}
\centering
\includegraphics[width=\linewidth]{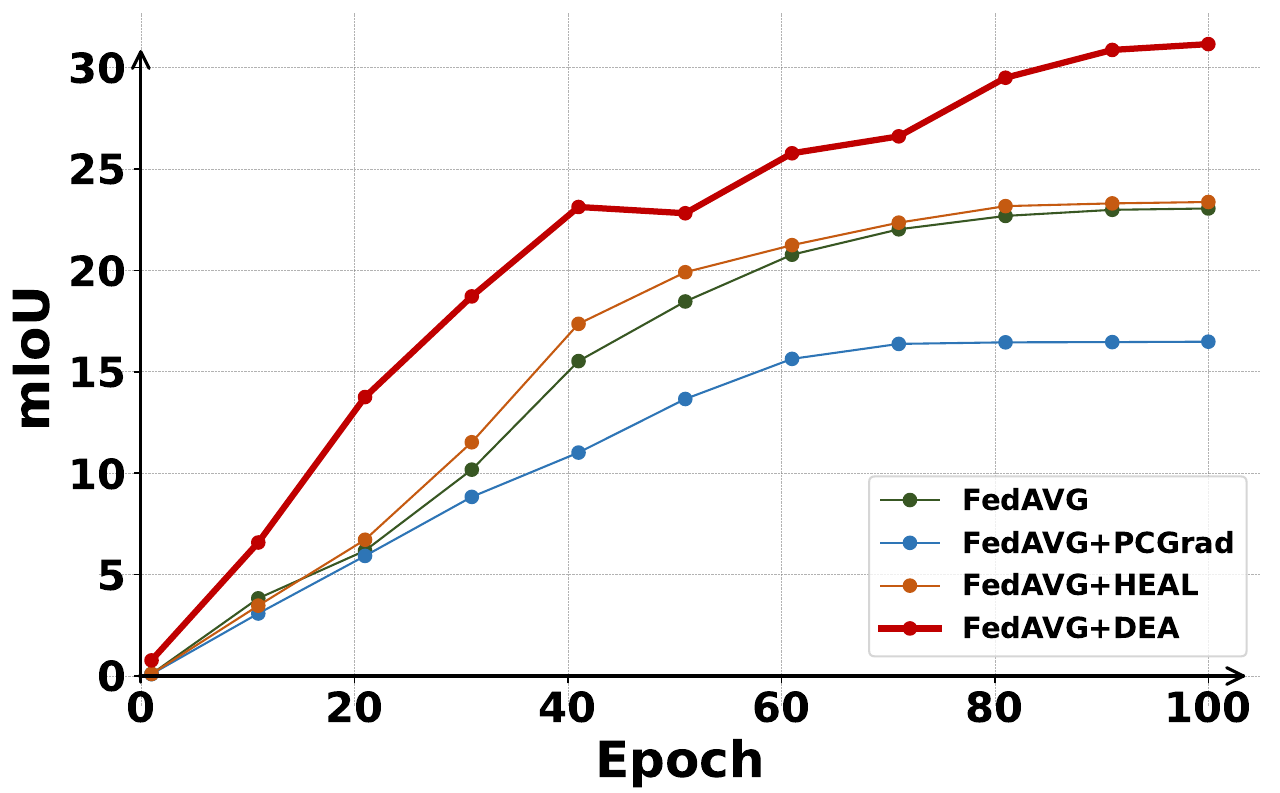}
\caption{Convergence comparison of FedAvg and its variants with PCGrad~\cite{pcgrad}, FedHEAL~\cite{fedhealCVPR2024}, and FedDEA.}
\label{fig:convergence-nyud-other}
\end{minipage}
\hfill
\begin{minipage}[t]{0.48\textwidth}
\centering
\includegraphics[width=\linewidth]{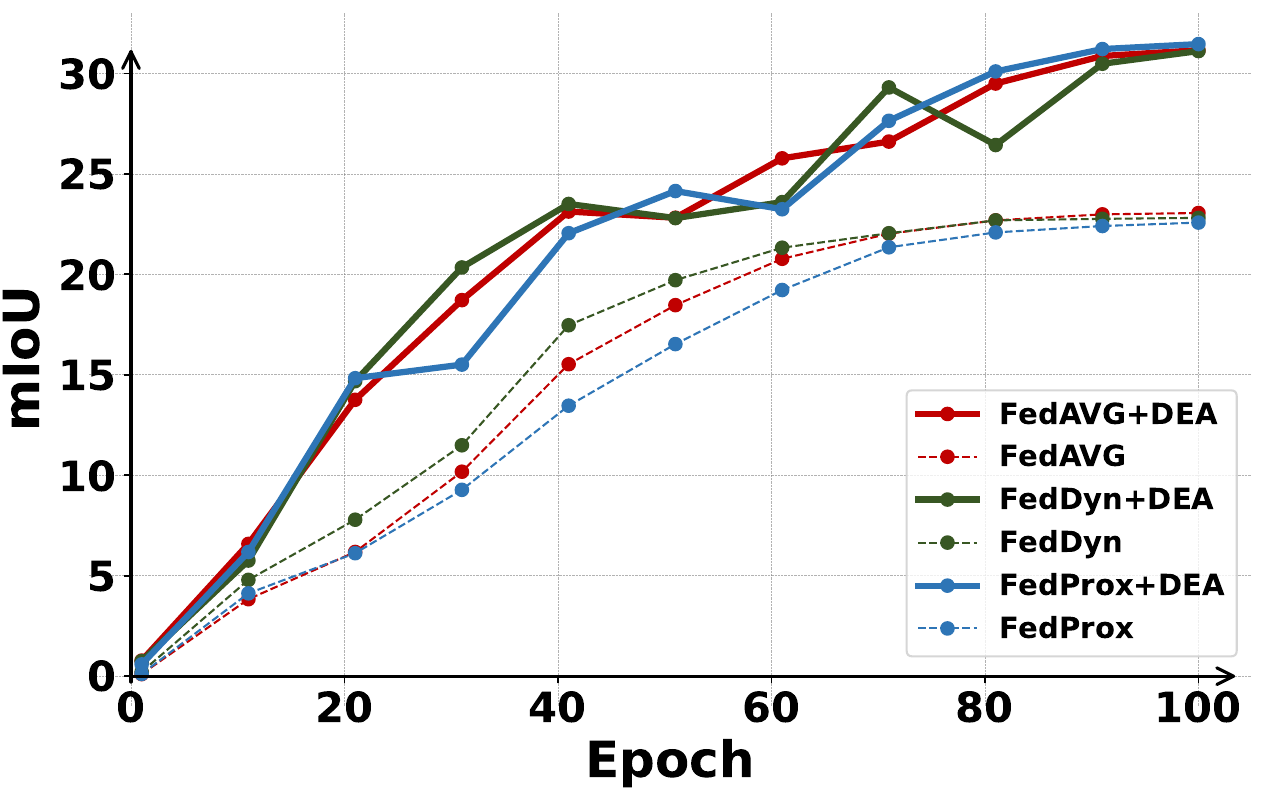}
\caption{Comparison of convergence trends for FedAvg, FedProx, and FedDyn after incorporating FedDEA .}
\label{fig:convergence-pascal-common}
\end{minipage}
\vspace{-12pt}
\end{figure}

\paragraph{Convergence Analysis.}
We conducted two sets of convergence experiments on the semantic segmentation task using the NYUD-V2. The first set uses FedAvg as the baseline and compares its convergence behavior after incorporating PCGrad~\cite{pcgrad}, FedHEAL~\cite{fedhealCVPR2024}, and FedDEA. The second set examines the training curves of FedAvg~\cite{fedavg2017}, FedProx~\cite{fedproxPMLS2020}, and FedDyn~\cite{feddynICLR2021} after integrating FedDEA. As shown in Figure~\ref{fig:convergence-nyud-other} and Figure~\ref{fig:convergence-pascal-common}, the results validate FedDEA's advantages in stabilizing training and improving convergence efficiency.

\section{Conclusion}
This paper focuses on a critical and practical problem in Federated Multi-Task Learning: \textbf{how to achieve unified modeling in the presence of task heterogeneity.} To address this challenge, we propose FedDEA, a 
update structure aware aggregation strategy designed to mitigate cross task update interference and enable collaborative modeling across heterogeneous tasks. FedDEA can be easily integrated as a plug-in-play module into mainstream federated optimization algorithms. Extensive empirical studies on two representative heterogeneous multi-task datasets NYUD-V2 and PASCAL-Context demonstrate that: (1) different tasks activate different regions in the parameter space; (2) decomposing the update space and suppressing interfering signals effectively alleviates parameter conflicts; and (3) FedDEA consistently improves performance across multiple federated methods, validating its robustness and generality under strong heterogeneity. In the future, we will extend this approach to multimodal scenarios to further enhance the practicality of federated multi-task learning.
 
\textbf{Potential Impact.} Existing federated learning methods struggle to resolve update interface in multi-task scenarios. FedDEA can leverage a masking mechanism to effectively mitigate task interference, and easily integrate into other federated learning frameworks. In the future, FedDEA is expected to leverage multi-center data from different tasks to train models supporting more functionalities, thereby better serving society.

\medskip
\newpage

{
\small
\bibliographystyle{plain}
\bibliography{reference}
}


\appendix
\newpage
\section{Datasets and Preprocessing Details}

We conduct experiments on two widely-used multi-task learning benchmarks: NYUD-V2~\cite{NYUD-V2} and PASCAL-Context~\cite{PASCAL-context}. Both datasets are processed following established practices in Multi-Task Learning~\cite{kokkinos2017ubernet,xu2018pad,vandenhende2020mti} and Federated Multi-Task Learning~\cite{MTFLCVPR2023, fedhca2CVPR2024} literature. Each task is treated as a distinct client in our federated setup, with isolated supervision and disjoint data partitions to simulate real-world task-heterogeneous scenarios. As for the two datasets, we split them using official setup as reported in the original papers~\cite{NYUD-V2,PASCAL-context}.

\subsection{NYUD-V2 Dataset}

\paragraph{Tasks:} Semantic segmentation, depth estimation, surface normal prediction, and edge detection. Output formats include single-channel maps (depth, edge), three-channel unit vectors (normals), and integer label masks (segmentation).

\paragraph{Preprocessing:} RGB images are resized to $448 \times 576$ and normalized using ImageNet~\cite{deng2009imagenet} statistics. Depth maps are scaled to a fixed range; semantic labels use $255$ as the ignore index; surface normals are normalized to $[-1, 1]$ and masked by magnitude; edge maps are derived via Laplacian filtering and binarization.

\subsection{PASCAL-Context Dataset}

\paragraph{Tasks:} Semantic segmentation, human part segmentation, surface normal prediction, saliency detection, and edge detection. Each task follows its own label schema.

\paragraph{Preprocessing:} Images are resized to $512 \times 512$. Segmentation labels are integer-encoded and padded. Saliency maps are binarized from grayscale with a $0.5$ threshold. Edge maps are extracted using Laplacian filters and refined by thinning. Normals are decoded from RGB to $[-1, 1]$ vectors and masked by valid semantic regions.

\section{Model Architectures}

\subsection{Main Model: Swin-T Encoder + FCN Decoder}

\paragraph{Encoder and decoder structure.}  
The primary model adopts a Swin-T~\cite{SwinTransformerCVPR2021} backbone as the shared encoder. It is configured with embedding dimensions of 96 and depths $(2, 2, 6, 2)$, and outputs four-stage hierarchical features with channel sizes of $[96, 192, 384, 768]$. The input image is tokenized by non-overlapping convolutional patch embeddings, followed by sequential processing through Swin Transformer blocks with local self-attention and patch merging. Outputs from all four stages are collected for decoding.

Each task is equipped with an independent decoder and prediction head. The decoder receives multi-level features and performs dimension reduction via linear projections. These features are then spatially aligned through upsampling to the same resolution and fused using $1\times1$ convolution and batch normalization, producing a task-specific feature map at $1/4$ resolution. This is followed by a two-stage transposed convolution head that upsamples the feature map back to the input resolution, followed by a $1\times1$ convolution to project to the task-specific number of output channels.

\paragraph{Parameter initialization and training setup.}  
The Swin-T~\cite{SwinTransformerCVPR2021} backbone is initialized from publicly available ImageNet-pretrained weights\cite{deng2009imagenet}. Decoders are initialized using He normal initialization~\cite{he2015delving}. Cross-entropy is used for classification tasks (e.g., segmentation, edge), L1 or MSE loss for regression tasks (e.g., depth, normals), and binary cross-entropy for saliency and edge detection.

\subsection{Extended Model: ResNet-18 Encoder + FCN Decoder}

\paragraph{Architectural differences from Swin-T setup.}  
To evaluate the robustness of our method under different encoder configurations, we additionally implement a variant using ResNet-18~\cite{he2016deep} as the encoder. This backbone is a $4$-stage convolutional network with channel dimensions $[64, 128, 256, 512]$, following standard residual block designs. Unlike the transformer-based Swin-T~\cite{SwinTransformerCVPR2021}, ResNet~\cite{he2016deep} uses spatial convolutions with downsampling via pooling and stride.

The decoders are reused with minimal adaptation. The decoder reshapes and projects the ResNet feature maps to a shared embedding dimension and fuses them identically as in the main Swin-based model.

\paragraph{Motivation for this extension.}  
The ResNet-18~\cite{he2016deep} variant serves as a lightweight and widely adopted baseline in computer vision tasks. Including this architecture allows us to evaluate the generalizability of our aggregation strategy across encoder families with distinct architectural inductive biases.

\begin{table*}[!t]
\caption{Performance comparison across NYUD~\cite{NYUD-V2} and PASCAL~\cite{PASCAL-context} datasets using different enhancement modules with ResNet-18~\cite{he2016deep} backbone. Evaluation metrics for each task follow the definitions provided in the main paper, and are consistent across both datasets. The symbol ``$\Delta$'' denotes the overall percentage improvement or degradation in performance relative to the corresponding base method, as defined in the main paper.
}
\centering
\small
\renewcommand\arraystretch{1.1}
\resizebox{\linewidth}{!}{
\begin{tabular}{r||cccc|c||ccccc|c}
\hline\thickhline
\rowcolor{lightgray}
& \multicolumn{5}{c||}{\textbf{NYUD}} & \multicolumn{6}{c}{\textbf{PASCAL}} \\
\cline{2-12}
\rowcolor{lightgray}
\multirow{-2}{*}{\textbf{Method}} 
& \makecell{Semseg \\ (mIoU)$\uparrow$} 
& \makecell{Depth \\ (RMSE)$\downarrow$} 
& \makecell{Normals \\ (mErr)$\downarrow$} 
& \makecell{Edge \\ (OdsF)$\uparrow$} 
& $\Delta$ 
& \makecell{Semseg \\ (mIoU)$\uparrow$} 
& \makecell{Parts \\ (mIoU)$\uparrow$} 
& \makecell{Normals \\ (mErr)$\downarrow$} 
& \makecell{Sal \\ (maxF)$\uparrow$} 
& \makecell{Edge \\ (OdsF)$\uparrow$} 
& $\Delta$ \\
\arrayrulecolor{gray!60}\Xhline{0.8pt}
\arrayrulecolor{black}
FedAvg\pub{AISTATS17}~\cite{fedavg2017}     & 7.29  & 0.8349  & 28.08  & 72.78  & -      & 13.93  & 34.10  & 18.51  & 75.40  & 67.15  & - \\
+PCGrad\pub{NeurIPS20}~\cite{pcgrad}          & 4.76  & 2.0593  & 38.97  & 68.02  & -56.67\% & 9.41   & 27.20  & 70.63  & 70.08  & 59.95  & -70.40\% \\
+HEAL\pub{CVPR24}~\cite{fedhealCVPR2024}            & 7.32  & 0.8540  & 28.17  & 72.54  & -0.62\%  & 14.50  & 30.26  & 18.58  & 75.51  & 67.51  & -1.37\% \\
\textbf{+DEA}    & 11.98 & 0.8494  & 27.84  & 72.74  & \textbf{+15.88\%} & 23.54  & 42.72  & 18.54  & 75.49  & 67.07  & \textbf{+18.82\%} \\
\hline
FedProx\pub{MLSys20}~\cite{fedproxPMLS2020}          & 6.22  & 0.8412  & 28.34  & 72.64  & -      & 11.45  & 33.74  & 18.60  & 75.55  & 67.20  & - \\
+PCGrad\pub{NeurIPS20}~\cite{pcgrad}          & 5.54  & 2.1193  & 35.61  & 66.92  & -49.11\% & 7.03   & 26.70  & 77.41  & 69.88  & 60.89  & -78.53\% \\
+HEAL\pub{CVPR24}~\cite{fedhealCVPR2024}            & 6.87  & 0.8388  & 28.07  & 64.52  & +0.14\%  & 10.84  & 30.66  & 18.70  & 75.73  & 67.23  & -2.95\% \\
\textbf{+DEA}    & 10.22 & 0.8526  & 27.98  & 72.37  & \textbf{+15.96\%} & 21.07  & 42.24  & 18.45  & 75.56  & 67.38  & \textbf{+22.06\%} \\
\hline
FedNova\pub{NeurIPS20}~\cite{fednovaNIPS2020}          & 6.38  & 0.8450  & 28.11  & 72.28  & -      & 14.48  & 35.79  & 18.62  & 75.51  & 67.11  & - \\
+PCGrad\pub{NeurIPS20}~\cite{pcgrad}          & 2.39  & 0.9023  & 29.03  & 72.15  & -18.20\% & 4.06   & 21.99  & 20.48  & 75.17  & 65.75  & -24.60\% \\
+HEAL\pub{CVPR24}~\cite{fedhealCVPR2024}            & 9.15  & 0.8418  & 28.03  & 72.60  & +6.16\%  & 11.54  & 39.01  & 18.69  & 75.52  & 66.78  & -2.44\% \\
\textbf{+DEA}    & 14.08 & 0.8420  & 27.68  & 72.56  & \textbf{+23.39\%} & 29.47  & 43.51  & 18.14  & 75.43  & 67.33  & \textbf{+25.56\%} \\
\hline
FedDyn\pub{ICLR21}~\cite{feddynICLR2021}           & 6.20  & 0.8429  & 27.99  & 72.38  & -      & 15.36  & 35.63  & 18.87  & 74.87  & 67.37  & - \\
+PCGrad\pub{NeurIPS20}~\cite{pcgrad}          & 4.60  & 2.1045  & 37.97  & 65.57  & -55.17\% & 8.49   & 26.10  & 70.67  & 69.13  & 59.32  & -73.12\% \\
+HEAL\pub{CVPR24}~\cite{fedhealCVPR2024}            & 7.74  & 0.8525  & 27.84  & 72.32  & +6.01\%  & 14.39  & 30.29  & 18.87  & 75.62  & 66.87  & -4.21\% \\
\textbf{+DEA}    & 11.70 & 0.8475  & 27.91  & 72.51  & \textbf{+22.11\%} & 27.05  & 42.27  & 18.37  & 75.34  & 67.98  & \textbf{+19.78\%} \\
\hline
Moon\pub{CVPR21}~\cite{moonCVPR2021}             & 5.50  & 0.8784  & 28.15  & 72.43  & -      & 7.95   & 24.79  & 21.77  & 74.73  & 66.66  & - \\
+PCGrad\pub{NeurIPS20}~\cite{pcgrad}          & 1.83  & 2.3325  & 42.72  & 61.76  & -74.68\% & 0.45   & 4.62   & 89.81  & 41.36  & 19.09  & -120.85\% \\
+HEAL\pub{CVPR24}~\cite{fedhealCVPR2024}            & 6.29  & 0.8482  & 28.44  & 72.32  & +4.14\%  & 12.12  & 33.05  & 18.75  & 75.63  & 63.19  & +19.13\% \\
\textbf{+DEA}    & 10.55 & 0.8491  & 27.74  & 72.51  & \textbf{+24.20\%} & 25.38  & 42.63  & 18.33  & 75.45  & 67.66  & \textbf{+61.91\%} \\
\hline
FedBN\pub{ICLR21}~\cite{fedbnICLR2021}             & 5.52  & 1.1024  & 32.25  & 69.37  & -      & 11.39  & 21.26  & 41.41  & 70.10  & 57.07  & - \\
+PCGrad\pub{NeurIPS20}~\cite{pcgrad}          & 2.68  & 1.3254  & 34.89  & 66.38  & -21.05\% & 1.71   & 2.69   & 87.15  & 21.19  & 15.26  & -85.17\% \\
+HEAL\pub{CVPR24}~\cite{fedhealCVPR2024}            & 5.14  & 1.0356  & 34.36  & 68.43  & -2.18\%  & 9.86   & 19.72  & 41.05  & 69.31  & 59.51  & -3.33\% \\
\textbf{+DEA}    & 6.78  & 1.0736  & 33.90  & 68.48  & \textbf{+4.75\%}  & 13.55  & 23.48  & 35.80  & 66.43  & 59.13  & \textbf{+8.27\%} \\
\hline
FedGA\pub{CVPR23}~\cite{fedgaCVPR2023}            & 8.95  & 0.8351  & 47.75  & 72.92  & -      & 18.31  & 34.50  & 18.57  & 75.84  & 67.17  & - \\
+PCGrad\pub{NeurIPS20}~\cite{pcgrad}          & 1.92  & 2.2631  & 48.81  & 67.29  & -64.88\% & 5.77   & 4.89   & 111.12 & 54.76  & 60.70  & -138.00\% \\
+HEAL\pub{CVPR24}~\cite{fedhealCVPR2024}            & 7.07  & 0.8461  & 48.29  & 72.51  & -6.03\%  & 16.65  & 31.36  & 18.39  & 75.76  & 67.01  & -3.51\% \\
\textbf{+DEA}    & 9.52  & 0.8433  & 47.21  & 72.56  & \textbf{+1.49\%}  & 22.55  & 39.48  & 18.48  & 75.30  & 67.33  & \textbf{+7.52\%} \\
\hline
FedACG\pub{CVPR24}~\cite{fedacgCVPR2024}           & 14.00 & 0.8495  & 28.10  & 72.21  & -      & 29.99  & 43.71  & 18.04  & 75.26  & 66.87  & - \\
+PCGrad\pub{NeurIPS20}~\cite{pcgrad}          & 13.46 & 0.8278  & 28.20  & 72.79  & -0.21\%  & 27.89  & 46.40  & 17.78  & 75.51  & 67.21  & 0.29\% \\
+HEAL\pub{CVPR24}~\cite{fedhealCVPR2024}            & 15.52 & 0.8461  & 28.10  & 72.87  & +3.06\%  & 28.51  & 43.59  & 18.13  & 75.42  & 67.22  & -0.99\% \\
\textbf{+DEA}    & 16.02 & 0.8406  & 28.05  & 72.75  & \textbf{+4.11\%}  & 33.97  & 44.37  & 17.90  & 75.46  & 67.87  & \textbf{+3.47\%} \\
\thickhline
\end{tabular}
}
\label{tab:rq2-enhancement}
\end{table*}

\section{Federated Optimization Settings}

Each federated experiment is conducted under a task-level partitioning, where the number of clients equals the number of tasks in the dataset. Specifically, the NYUD-V2~\cite{NYUD-V2} benchmark contains four tasks, resulting in four clients, and the PASCAL-Context~\cite{PASCAL-context} benchmark contains five tasks, resulting in five clients. 

To account for the disparity in dataset size between NYUD-V2~\cite{NYUD-V2} and PASCAL-Context~\cite{PASCAL-context}, we adjust the number of local training epochs per round accordingly. For NYUD-V2~\cite{NYUD-V2}, each client trains for $5$ local epochs per communication round, while for PASCAL-Context~\cite{PASCAL-context}, the local epoch count is set to $2$. The batch size for training is set to $8$, and the validation batch size is set to $4$. 

All experiments are conducted for a total of 100 communication rounds; model synchronization occurs at the end of every round. We adopt the AdamW optimizer for all clients, with a fixed initial learning rate of $1 \times 10^{-4}$, weight decay of $1 \times 10^{-4}$, and a linear warmup strategy during the first 5 local epochs. After the warmup phase, we apply a cosine annealing scheduler to progressively reduce the learning rate, enabling smoother convergence dynamics.

\section{Evaluation with ResNet-18 Backbone}
\subsection{Generalizability Across Different Aggregation Strategies under ResNet-18}

To further assess the robustness of FedDEA, we evaluate its performance under diverse federated optimization strategies using a lightweight ResNet-18~\cite{he2016deep} backbone. Specifically, we compare FedDEA with PCGrad~\cite{pcgrad} and FedHeal~\cite{fedhealCVPR2024} when applied to FedAvg~\cite{fedavg2017}, FedProx~\cite{fedproxPMLS2020}, FedNova~\cite{fednovaNIPS2020}, Moon~\cite{moonCVPR2021}, FedDyn~\cite{feddynICLR2021}, FedBN~\cite{fedbnICLR2021}, FedGA~\cite{fedgaCVPR2023}, and FedACG~\cite{fedacgCVPR2024}, using the same training setup and task partitioning as in the Swin-T~\cite{SwinTransformerCVPR2021} experiments. Results are shown in Table~\ref{tab:rq2-enhancement}

\begin{itemize}
    \item \textbf{Obs1:} Across all tested strategies, FedDEA maintains consistently competitive performance under ResNet-18~\cite{he2016deep}, reinforcing that the method's gains are not limited to a specific optimization algorithm. This observation aligns with our findings in the main paper, demonstrating that the proposed disentanglement and aggregation approach is broadly compatible with mainstream FL optimizers.
    
    \item \textbf{Obs2:} The ranking of strategies under ResNet-18~\cite{he2016deep} mirrors that observed with Swin-T~\cite{SwinTransformerCVPR2021}.  This consistency suggests that our framework generalizes well across heterogeneous aggregation settings, regardless of backbone complexity.
    
    \item \textbf{Obs3:} Based on the conclusions drawn in the main paper and the results under the ResNet-18 backbone, FedDEA proves to be both baseline-agnostic and architecture-agnostic. Its consistent superiority across diverse settings highlights that the strength of our approach stems fundamentally from the principled assumptions we make about the effectiveness of parameter updates.
\end{itemize}

\begin{table}[t]
\centering
\caption{Ablation study of masking and rescaling strategies on NYUD-V2 dataset.}
\label{tab:glad-ablation-nyud}
\renewcommand\arraystretch{1.1}
\begin{adjustbox}{max width=\textwidth}
\begin{tabular}{l||cccc|c}
\hline\thickhline
\rowcolor{lightgray}
& \multicolumn{5}{c}{\textbf{NYUD}} \\
\cline{2-6}
\rowcolor{lightgray}
\multirow{-2}{*}{\textbf{Method}} 
& \makecell{Semseg \\ (mIoU)$\uparrow$} 
& \makecell{Depth \\ (RMSE)$\downarrow$} 
& \makecell{Normals \\ (mErr)$\downarrow$} 
& \makecell{Edge \\ (OdsF)$\uparrow$} 
& $\Delta$\% \\
\arrayrulecolor{gray!60}\Xhline{0.8pt}
\arrayrulecolor{black}
FedAvg            & 7.29  & 0.8349  & 28.08  & 72.78  & --        \\
+DEA(a)       & 2.92  & 1.2037  & 34.55  & 69.68  & -32.84\%  \\
+DEA(b)        & 6.73  & 0.8428  & 27.98  & 72.41  & -2.17\%   \\
+DEA(c)       & 7.62  & 0.8567  & 28.12  & 72.90  & +0.50\%   \\
\textbf{+DEA}     & 11.98 & 0.8494 & 27.84 & 72.74 & \textbf{+15.88\%} \\
\thickhline
\end{tabular}
\end{adjustbox}
\vspace{-1ex}
\end{table}

\subsection{Methodological Analysis with Lightweight Encoder}

\paragraph{Ablation Analysis of Decoupling and Recalibration Mechanisms.}
To systematically assess the contribution of FedDEA's core components under a lightweight encoder, we design three controlled variants targeting its masking and gradient rescaling strategies: \textbf{(a) Small Mask}, which retains only parameter dimensions with relatively small gradient magnitudes; \textbf{(b) No Rescale}, which omits the rescaling step entirely; and \textbf{(c) Random Mask}, which randomly selects a subset of dimensions to retain. As shown in Table~\ref{tab:glad-ablation-nyud}, the Small Mask variant yields a significant drop in performance, indicating that dimensions with weak gradient signals tend to lack informative structure. Interestingly, the Random Mask variant performs comparably to the FedAvg baseline, suggesting that a large portion of parameter updates may be redundant or noisy. In contrast, the full FedDEA configuration --- leveraging structure-aware masking and calibrated rescaling --- achieves substantial improvements. These findings collectively demonstrate that both decoupling and recalibration are critical to FedDEA's effectiveness, even when deployed on a capacity-constrained backbone such as ResNet-18~\cite{he2016deep}.

\paragraph{Convergence Analysis.} To assess the stability and efficiency of DEA training under limited model capacity, we analyze its convergence behavior using a ResNet-18~\cite{he2016deep} backbone. Figure~\ref{fig:convergence-nyud-other} compares FedDEA with two representative generic modules, PCGrad~\cite{pcgrad} and FedHEAL~\cite{fedhealCVPR2024}, when integrated into the FedAvg~\cite{fedavg2017} framework. FedDEA demonstrates significantly smoother and more stable convergence, avoiding the early saturation and fluctuations observed in the baselines, thereby highlighting the advantages of our decoupled training strategy. These findings are consistent with the results reported in the main paper under the Swin-T~\cite{SwinTransformerCVPR2021} backbone, further confirming the convergence efficiency and optimization stability of FedDEA across different model capacities.

Figure~\ref{fig:convergence-pascal-common} further evaluates the plug-and-play compatibility of FedDEA by applying it to three federated optimizers: FedAvg~\cite{fedavg2017}, FedProx~\cite{fedproxPMLS2020}, and FedDyn~\cite{feddynICLR2021}. In all cases, FedDEA consistently improves convergence speed and stability, demonstrating strong adaptability and robustness. These results are also in line with the observations made under the Swin-T~\cite{SwinTransformerCVPR2021} architecture, further validating that FedDEA delivers consistently favorable convergence dynamics across diverse optimization frameworks.

\begin{figure}[t]
\centering
\begin{minipage}[t]{0.48\textwidth}
\centering
\includegraphics[width=\linewidth]{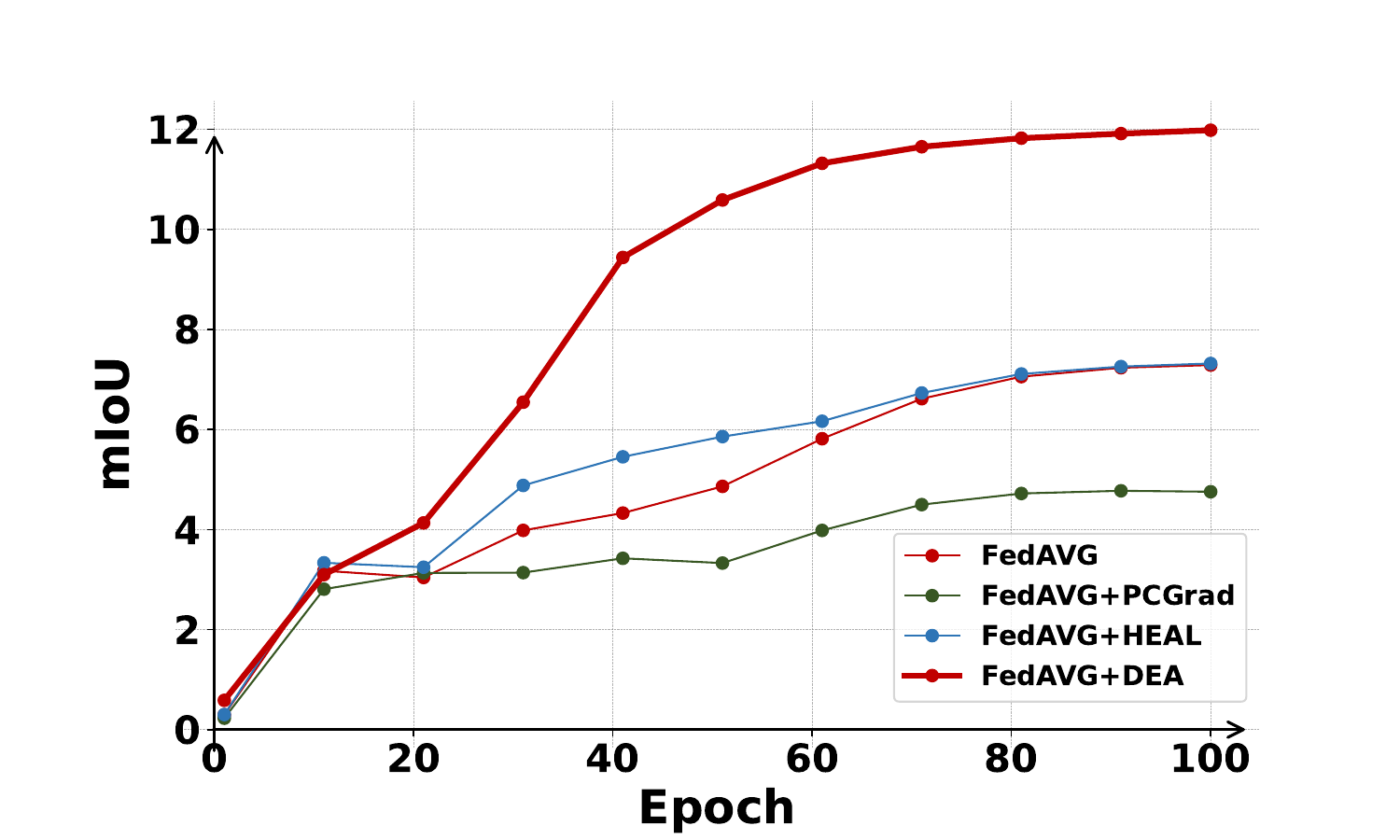}
\caption{Convergence comparison of FedAvg~\cite{fedavg2017} and its variants with PCGrad~\cite{pcgrad}, FedHEAL~\cite{fedhealCVPR2024}, and FedDEA.}
\label{fig:convergence-nyud-other}
\end{minipage}
\hfill
\begin{minipage}[t]{0.48\textwidth}
\centering
\includegraphics[width=\linewidth]{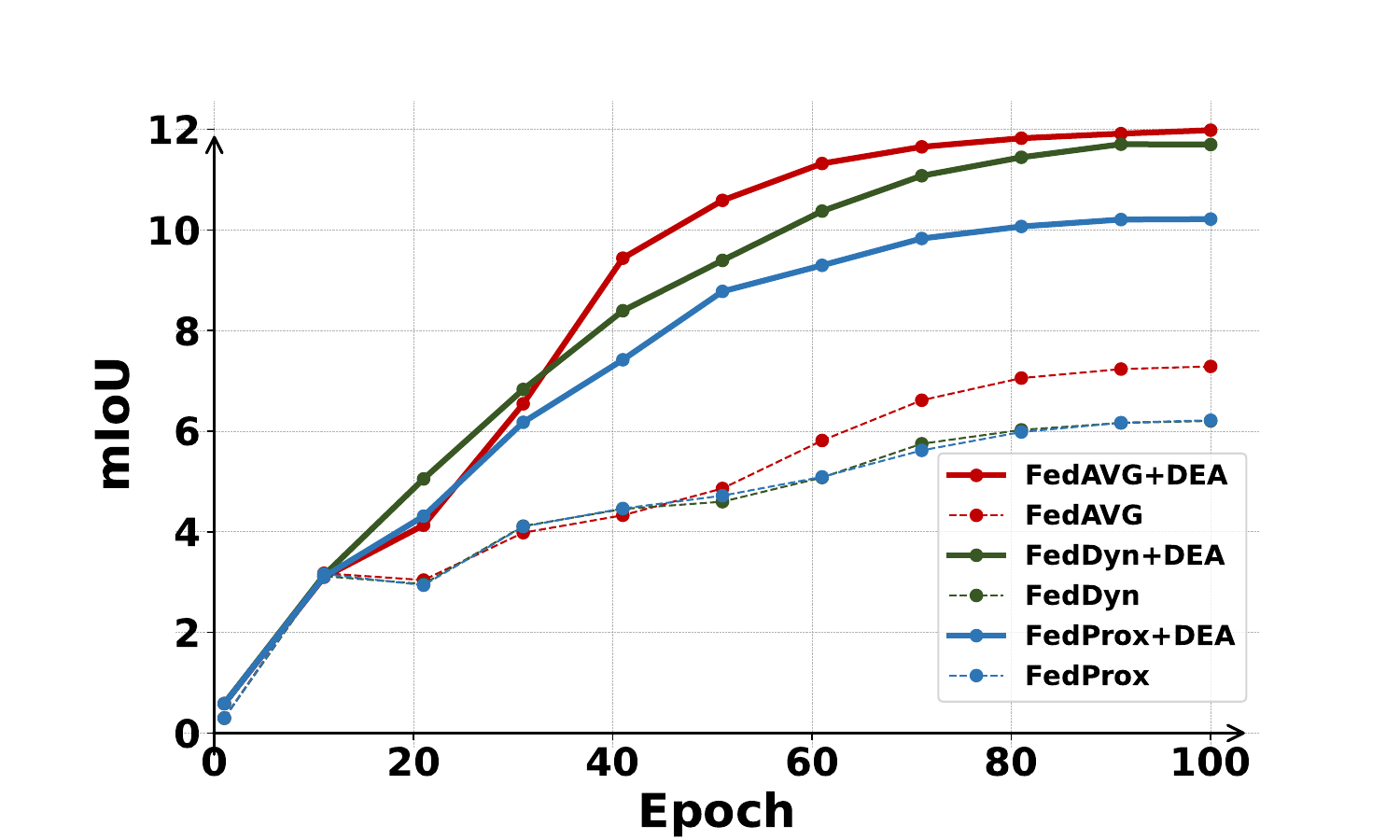}
\caption{Comparison of convergence trends for FedAvg~\cite{fedavg2017}, FedProx~\cite{fedproxPMLS2020}, and FedDyn~\cite{feddynICLR2021} after incorporating FedDEA .}
\label{fig:convergence-pascal-common}
\end{minipage}
\vspace{-12pt}
\end{figure}

\section{Compute and Reproducibility Details}

\subsection{Compute Resources}

All experiments were performed on a local research server equipped with eight NVIDIA RTX $3090$ GPUs. Specifically, the complete training process for any individual method on either dataset can be executed on a single RTX $3090$ GPU without memory issues.

On average, training a single method on the NYUD-V2~\cite{NYUD-V2} dataset requires approximately $8$ hours, while experiments on the PASCAL-Context~\cite{PASCAL-context} dataset take about $30$ hours. These averages are measured across multiple methods and reflect the total runtime for $100$ communication rounds, including both local client updates and server-side aggregation.

\subsection{Reproducibility}

To support reproducibility, we have submitted the complete experimental results for the Swin-T~\cite{SwinTransformerCVPR2021} backbone with DEA on the NYUD-V2~\cite{NYUD-V2} dataset. All hyperparameter settings, evaluation protocols, and implementation details are consistent with the descriptions in the main paper and supplementary sections.

Our codebase, along with configuration files and model weights, will be publicly released upon acceptance.
\newpage


\end{document}